\documentclass{article}
\usepackage{spconf,amsmath,graphicx}

\usepackage{multirow}
\usepackage{xspace}
\usepackage{url}
\newcommand*{\eg}{\emph{e.g.}\@\xspace}
\newcommand*{\ie}{\emph{i.e.}\@\xspace}

\newcommand*{\etc}{\emph{etc.}\@\xspace}

\newcommand*{\etal}{\emph{et al.}\@\xspace}

\title{TS-Net: Combining Modality Specific and Common Features\\ for Multimodal Patch Matching}
\name{Sovann En, Alexis Lechervy, Fr\'ed\'eric Jurie}
\address{Normandie Univ, UNICAEN, ENSICAEN, CNRS -- UMR GREYC}

\usepackage[symbol]{footmisc}
\renewcommand{\thefootnote}{\fnsymbol{footnote}}

\begin{document}
\maketitle
\renewcommand{\thefootnote}{\fnsymbol{footnote}}

\begin{abstract}
Multimodal patch matching addresses the problem of finding the correspondences between image patches from two different modalities, \eg RGB vs sketch or RGB vs near-infrared. The comparison of patches of different modalities can be done by discovering the information common to both modalities (Siamese like approaches) or the modality-specific information (Pseudo-Siamese like approaches). We observed that none of these two scenarios is optimal. This motivates us to propose a three-stream architecture, dubbed as TS-Net, combining the benefits of the two.  In addition, we show that adding extra constraints in the intermediate layers of such networks further boosts the performance. Experimentations on three multimodal datasets show significant performance gains in comparison with Siamese and Pseudo-Siamese networks\footnote[2]{Codes and resources available at \url{http://github.com/ensv/TS-Net}}.
\end{abstract}

\section{Introduction and related work}
\label{sec:intro}

Patch matching, the task consisting in determining the correspondences between image patches, is  essential for many computer vision problems, \ie, multi-view reconstruction, structure from motion, object-instance recognition, \etc In this work, we aim to study the problem of matching patches in a multimodal setting where input patches come from different sources, \ie RGB images vs hand-drawn sketches or RGB vs near-infrared images. 

% \begin{figure}
% \centering
% \includegraphics[width=0.47\textwidth]{images/introduction}
% \caption{Multi-modality patch-based matching to find corresponding RGB image from a database with the help of a partially drawn sketch.}
% \label{criminal}
% \end{figure}

Broadly speaking, there are two main ways to design local patch matching systems, either by employing hand-crafted features or through machine learning techniques. Pioneer works in patch matching \cite{lowe1999object} are based on handcrafted features  such as the SIFT descriptor/detector or some variants, \eg \cite{bay2006surf}, DAISY, \cite{tola2010daisy}, \etc Such approaches usually use conventional distance to measure patch similarity, \eg the Euclidean distance, which usually does not provide an optimal solution for matching purposes. This family of approaches relies heavily on human expertise. 

In contrast with feature engineering, another approach to patch matching consists in using supervised algorithms to find adapted features or adapted similarity functions, for given datasets. Machine learning allows to find optimal projections minimizing (or maximizing) the distances between  positive patches (negative patches respectively) \cite{heinly2012comparative,jain2012metric, brown2011discriminative}. 

Recent breakthroughs in deep learning have strongly contributed to this field. One of the first works in deep metric learning is  the one of Jahrer \etal \cite{jahrer2008learned} introducing a Siamese networks inspired by the LeNet5 networks,  and comparing the so-obtained features with the Euclidean distance. Since then, Siamese networks have been very popular in the literature. Several variants have been proposed, differing by their weight-sharing strategy \cite{zagoruyko2015learning} (Siamese vs Pseudo Siamese), combinations of the inputs \cite{zagoruyko2015learning,tian2017l2} (two channels input images vs multi-scale images), similarity functions (conventional distance \cite{zagoruyko2015learning, zbontar2016stereo,tian2017l2, altwaijry2016learning} or using metric layers \cite{han2015matchnet}).

Another important aspect when training deep networks for patch matching is the objective function. It can be (a) the cross entropy (binary classification loss)~\cite{han2015matchnet}  (b) the hinge loss~\cite{zagoruyko2015learning} (c) the triplet loss \cite{balntas2016pn,kumar2016learning} to incorporate the notion of relative distance, relative distance \cite{tian2017l2} (d) the global loss which models the loss as two distributions (positive and negative) to be pushed away from each other \cite{kumar2016learning}.

More specifically, the question of multimodal patch matching has been investigated recently by several authors. \cite{aguilera2016learning, suarez2017cross} suggested to concatenate the different modalities as different channels of the input data. \cite{merkle2017exploiting} experimented the use of Siamese networks for the matching of visible/SAR patches. 
\cite{aguilera2017cross} studied the quadratic network, a variant of the Siamese network that takes 4 patches as input. In the context of cross-spectral face recognition~\cite{lezama2017not} proposed two components (one before and another one after the feature extraction network) to allow the system to transform the NIR images into the VIS spectrum. %Interestingly, they show that using any one among the two produces effectively similar performance. 
As we write, Siamese networks are still seen as a reference for multimodal patch matching.

%As aforementioned, deep learning based methods for patch matching usually rely on Siamese architectures or their variants. They takes the input patches and projects them through several shared CNN layers into a new subspace where the distance can be directly computed (\ie using L2 distance) \cite{tian2017l2, simo2015discriminative} or by means of a metric network using FC layers \cite{han2015matchnet, kumar2016learning} (see Figure \ref{fig:network_architecture}). The system relies on L2 distance and assumes that the output of the network lies in the same Euclidean subspace and is linearly separable. However, this is not the case always, especially when it comes to a multimodal setting. That is the reason why we opt for a Siamese network with metric layers.

\begin{figure*}[tb]
\centering
\includegraphics[width=0.85\textwidth]{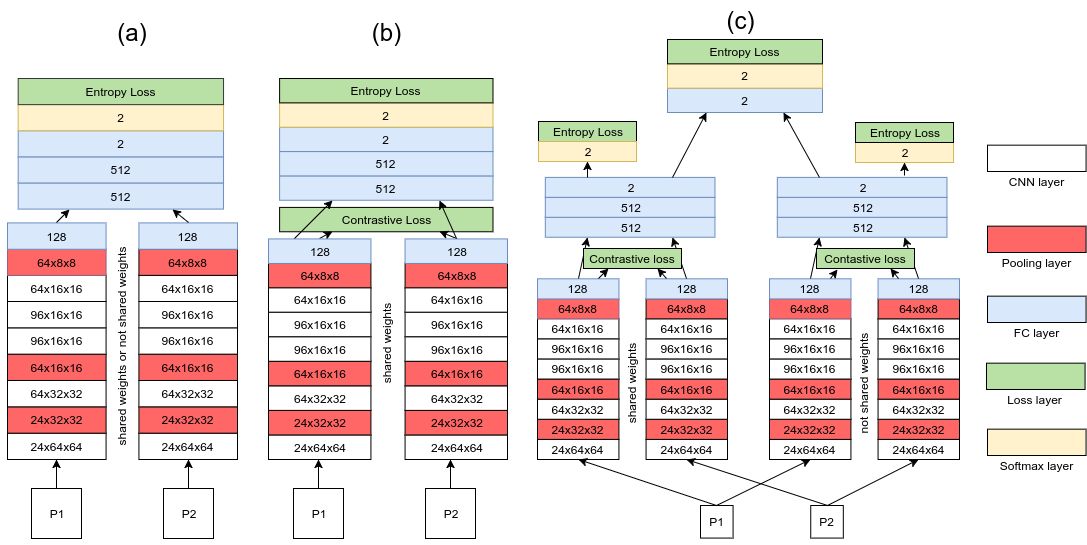}
\caption{The detailed architectures of (a) standard Siamese networks (b) Siamese networks with the proposed additional loss on the feature towers  (c) the proposed TS-Net network with additional losses on the feature extraction tower and on the metric network. The numbers on each rectangle indicate the output size of this layer.}
\label{fig:network_architecture}
\end{figure*}

One important aspect of  Siamese architectures is that the weights of feature extraction towers are shared between the inputs. This is to say that the network is trained to extract characteristics present in both modalities. In case of the Pseudo-Siamese architectures, the feature towers are not shared: contrarily to the Siamese networks, the motivation is to extract modality specific information in order to better discriminate the pair of inputs. Our motivation in this paper is to take advantage of these two complementary aspects and propose a novel architecture, dubbed as TS-Net. It consists of two sub-networks, one Siamese and one Pseudo-Siamese networks. Their outputs are  combined with a fully connected layer, acting as a weighting scheme between the modality specific information and the common information present in the input patches. The overall architecture is given Figure \ref{fig:network_architecture}.

Our second contribution is to show that adding a constraint on the feature embedding, by  means of a contrastive loss in the feature extraction tower, helps to boost the performance further. The idea is to encourage the network to bring  projections of positive pairs closer in the Euclidean space. In the extreme case, this is equivalent to make two clusters of input pairs at the metric layers, allowing to easily separate them with an hyperplane instead of having to learn an arbitrarily complex boundary.  

The rest of the paper is organized as follows:  Section \ref{sec:network_architecture} introduces  the network architecture and the training methodology. Section \ref{sec:experimentations} discusses the datasets  and presents the experimental validation of the approach. Finally, Section \ref{sec:conclusion} concludes the paper.

%\section{Related Work}
%\label{sec:related_work}

\section{The Proposed Three-Stream Network}
\label{sec:network_architecture}

As explained before, the proposed architecture for multimodal patch matching, denoted as the TS-Net architecture, is intended to combine the advantages of both Siamese and Pseudo-Siamese networks. The overall architecture of TS-Net  is given Fig. \ref{fig:network_architecture}(c). Each sub-network has 2 main parts: two feature extraction towers and a metric learning module. In the case of the Siamese network, the parameters of the feature extraction towers are shared, while for Pseudo-Siamese networks they are distinct.  TS-Net takes a pair of patches as input, one from each modality, and predicts independently in each sub-network whether they are similar or not. Finally, the outputs of each sub-network are combined by an additional fully connected (FC) layer to produce the final prediction. 
In the next paragraphs, the different components of TS-Net are described and commented.

\vspace{.8em}\noindent \textbf{Feature extraction network.} Each tower is based on convolutional and pooling layers to hierarchically extract high-level information from the input patches. We use max-pooling layers to reduce the dimensions of the feature maps by a factor of 2. At the end of the tower, we use a bottleneck (fully connected) layer to produce a compact output vector with 128 dimensions. Inspired from \cite{han2015matchnet}, we use $Relu$ activation as a non-linear activation function.

\vspace{.8em}\noindent \textbf{Tower Fusion.} We observed in our experiments that 
subtracting the layers produced better performance than concatenating them, as in the original MatchNet. So the output of the feature extraction tower are element-wise subtracted before they are fed to the metric network.

\vspace{.8em}\noindent \textbf{Metric network.} The metric learning part of the network consists of three fully connected layers. It takes a vector of 128 dimensions and produces a vector of dimension two, suitable for binary classification.

\vspace{.8em}\noindent \textbf{Losses.}  We treat patch matching as a binary classification problem, as we observed it performs better (also observed by \cite{tian2017l2}) than learning a similarity function. Therefore, Siamese and Pseudo-Siamese parts of TS-Net are trained with binary cross-entropy loss functions.

One contribution of this paper is to introduce additional constraints, at the feature level, by  means of a contrastive loss \cite{chopra2005learning} enforcing the features coming from the two feature towers to be close to each other if the pair is positive. This will enable the features of positive pairs to be in the hypersphere and the features of the negative pairs to be outside the hypersphere.

The fusion of Siamese and Pseudo-Siamese networks is done by introducing an additional cross-entropy loss on the top of the two. 

More formally, let $(x_1, x_2)$ be the input pair of patches and $y$ the class label. $y=1$ means the pair is positive (similar patches), $y=0$ means the pair is negative (different patches). We denote by $L_{en}$ and $L_{con}$  the cross-entropy and the contrastive loss, with:\\ (a) $L_{en} = y \log(\hat{y}) + (1-y)\log(1-\hat{y})$ where $\hat{y}$ is output of the \textit{Softmax} layer, and \\ (b) 
$L_{con} = y \frac{2}{Q}D^2 + (1-y)2Qe^{\frac{-2.77}{Q}D}$
%\label{equ:lambda}
%\end{equation}
where $D$ is the Euclidean distance between features. Q is the margin to be optimized.
The overall loss function is then given by:\\ $
L = L_{tsnet_{en}} + L_{siam_{en}} +   L_{pseudo_{en}} + \lambda L_{siam_{con}} +   \beta L_{pseudo_{con}}$, with $\lambda$ and $\beta $ two cross-validated parameters in $[0,1]$. 

In multimodal settings, it is not always guaranteed that the two modalities can be projected into the same subspace.  In practice, we observed that optimal performance is obtained for $\lambda$ and $\beta$  set to $10^{-2}$ (values obtained by cross validating the parameters on the validation set).

\vspace{.8em}\noindent \textbf{Implementation details.} We initialize the weights of each convolutional layer  using the Xavier initialization and all the FC layers with a truncated normal distribution ($stddev=0.005$ and $mean=0$, $bias=0.1$). While the original MatchNet is trained with plain stochastic gradient descent, we found that training with 0.95 momentum produce equal or better performance. We train the network with $lr=10^{-3}$ with L2 regularization of $10^{-3}$ with neither dropout nor $batchnorm$. $Q$ is optimized experimentally on VeDAI validation set and set to be 50 for the other two datasets. During training, we observe that the $\lambda$ and $\beta$ parameters should be carefully set and the best performance we obtain is for $\lambda = 10^{-2}, \beta = 10^{-4}$ or $\beta = 10^{-2}$ on CUHK and NIR Scene (cross validation experiments). We use batch size of 32 and train with at least 150 epochs. All the experimentations are done using Tensorflow 1.4 with NVIDIA P100 or K80 GPU. Patches are normalized to have zero mean and unit standard deviation for each modality.
\section{Experimentations}
\label{sec:experimentations}
Our aim in this section is to provide insights about TS-Net, its behavior and, more importantly, to draw a comparison with Siamese and Pseudo-Siamese networks, which are considered as a reference to this task. First, we run a series of experiments on the VeDAI dataset to validate TS-Net. It consists in evaluating different ways to fuse information either in the metric or after the feature extraction network. Next, we show that the gain in performance is not due to an increase of the number of parameters. Finally, we run experiments on  three public datasets to experimentally validate our network and compare it to Siamese and Pseudo-Siamese networks. To report the performance, we employ the standard evaluation protocol defined in \cite{brown2011discriminative}, namely the {\em 95\% error rate} criteria, abbreviated 95\%ErrRate, which is the percentage of false matches present when 95\% of all correct matches are detected. For each experimentation, we report the average performance with its standard deviation on at least 3 runs (Table \ref{tab:tsnet_fusion}) and 8 runs (Table \ref{tab:multi_matching_results}). 

\vspace{.8em}\noindent \textbf{Datasets} The proposed approach is experimentally validated on three different datasets: VeDAI 
\footnote{\url{https://downloads.greyc.fr/vedai/}}, RGB-NIR Scene \footnote{\url{https://ivrl.epfl.ch/supplementary_material/cvpr11/}} and CUHK \footnote{\url{http://mmlab.ie.cuhk.edu.hk/archive/facesketch.html}}. These 3 datasets contain images from two different modalities. It is worth mentioning that these 3 datasets have been created for different tasks. Therefore, it will provide an opportunity to test and compare performance on a variety of fields. For instance, VeDAI  is generally used for Vehicle Detection in Aerial Imagery while CUHK for face sketch synthesis/recognition. VeDAI, CUHK and RGB-NIR Scene  contain respectively a total of 1246, 188 and 477 pairs of images.

\vspace{.8em}\noindent \textbf{Pairs of Patch Generation.} For each dataset, the images are given as sets of aligned pairs (one image from each modality). To extract patches and form pairs, we uniformly sample each image using grid-like layout where each cell has a width and height of 64 $\times$ 64 pixels. This gives us a collection of corresponding positive patches. We randomly choose patches coming from different images to form negative pairs.

To make our patch matching experiments more realistic and challenging, we artificially augment our datasets by introducing some random affine transformations between the images of the same pair. For each pair, we generate three additional pairs using a random combination of: (i) Rotation (-12 to 12 degrees), (ii) Translation (-5 to 5 pixels on both axes) and (iii)  Scale (0.8 to 0.99). For the validation and test set, we  keep only one pair among the four, chosen randomly. Table \ref{tab:dataset} summarizes the number of train, test and validation pairs of patches. Half are positive, half are negative.

\begin{table}
\centering
\caption{Number of pairs of patches in the train, test and validation set, for each dataset. Each set contains 50\% of positive pairs and 50\% of negative ones.}
\begin{tabular}{llll}
Dataset   & Train (70\%) & Test (20\%) & Validation (10\%) \\ \hline
VeDAI     & 448k               & 128k              & 64k                     \\
CUHK      & 113k               & 32k               & 16k                     \\
NIR Scene & 427k               & 122k              & 61k           
\end{tabular}
\label{tab:dataset}
\vspace{-0.8em}
\end{table}

\begin{table}
\centering
\vspace{-1em}\caption{95\%ErrRate on VeDAI validation set using TS-Net. Rows: tower fusion after the feature extraction network (bottleneck layer), FC1, FC2 or FC3 of the metric layer. `1 Entropy` means there is only one classification loss at the top of the network. `3 Entropy`:  each sub-network also has his own classification loss. S*:   Matchnet Network with the same number of parameters as TS-Net.}\label{tab:tsnet_fusion}
\begin{tabular}{lll}
              & 3 Entropy losses         & 1 Entropy loss \\ \hline
FC3 (TS-Net)   & \textbf{0.52 $\pm$ 0.07 }   & 0.93 $\pm$ 0.05 \\
FC2           & 0.62 $\pm$ 0.13             & 0.92 $\pm$ 0.05 \\
FC1           & 0.74 $\pm$ 0.07             & 1.03 $\pm$ 0.06 \\
Feature tower & n/a                         & 1.05 $\pm$ 0.07 \\
S*     & n/a                         & 1.01 $\pm$ 0.11
\end{tabular}\vspace{-1em}
\end{table}

\vspace{.8em}\noindent \textbf{Combining  Siamese and  Pseudo-Siamese networks.} Our motivation is to find an efficient way to combine the information coming from the two sub-networks. We consider four options depending on whether this fusion (element-wise subtraction) is done (a) after the feature extraction tower (b) after the first (c) second or (d) third layer of the metric network. In the case of early fusion, all the following layers are kept as in MatchNet. Table \ref{tab:tsnet_fusion} shows the performance given by each alternative. It also compares the performance obtained when 1 unique entropy loss ($tsnet_{en}$) is used, on the top of the network, with the performance obtained when each sub-network has, in addition, its own loss ($L_{tsnet_{en}} + L_{siam_{en}} + L_{pseudo_{en}}$). Based on these results, it is clear that the additional losses are important. The two additional losses help to guarantee the Siamese and the Pseudo-Siamese network learn complementary representation of the modalities. Consequently, this is the reason why having a late fusion (after FC3) is more beneficial. In addition, to guarantee that the gain in performance of TS-Net is not due to a larger number of parameters, we also provide the performance of  MatchNet (noted as S* in Table \ref{tab:tsnet_fusion}) when we increase the number of parameters in the feature tower by a factor of 1.45 and the bottleneck by 2 to have exactly the same number of parameters as in TS-Net. Experimental results suggest that this is roughly equivalent to the performance of TS-Net without additional losses with fusion at the FC1 layer.

\begin{table}[tb]
\centering
\caption{95\%ErrRate  on the 3 datasets, for Siamese network alone (S), Pseudo-Siamese network alone (PS), TS-NET, without/with the additional contrastive loss (C).}
\label{tab:multi_matching_results}
\begin{tabular}{llll}
% Dataset & Vedai           & CUHK            & NIR Scene        \\ \hline
% S       & 1.07 $\pm$ 0.07 & 5.38 $\pm$ 0.38 & 14.47 $\pm$ 0.53 \\
% PS      & 1.54 $\pm$ 0.05 & 5.63 $\pm$ 0.24 & 16.53 $\pm$ 0.61 \\
% TS-Net   & 0.56 $\pm$ 0.05 & 3.55 $\pm$ 0.06 & 12.55 $\pm$ 0.90 \\ \hline
% S+C     & 0.83 $\pm$ 0.01 & 3.34 $\pm$ 0.22 & 12.98 $\pm$ 0.22 \\
% PS+C    & 1.28 $\pm$ 0.03 & 3.80 $\pm$ 0.15 & 15.58 $\pm$ 0.34                  \\
% TS-Net+C & \textbf{0.53 $\pm$ 0.04}& \textbf{2.76 $\pm$ 0.08} & \textbf{11.47 $\pm$ 0.40}
Dataset & Vedai           & CUHK            & NIR Scene        \\ \hline
S       & 1.16 $\pm$ 0.07 & 5.07 $\pm$ 0.46 & 14.35 $\pm$ 0.20 \\
PS      & 1.50 $\pm$ 0.08 & 5.56 $\pm$ 0.36 & 16.05 $\pm$ 0.30 \\
TS-Net   & 0.52 $\pm$ 0.07 & 3.58 $\pm$ 0.14 & 12.40 $\pm$ 0.34 \\ \hline
S+C     & 0.84 $\pm$ 0.05 & 3.38 $\pm$ 0.20 & 13.17 $\pm$ 0.86 \\
PS+C    & 1.37 $\pm$ 0.08 & 3.70 $\pm$ 0.14 & 15.60 $\pm$ 0.28                  \\
TS-Net+C & \textbf{0.45 $\pm$ 0.05}& \textbf{2.77 $\pm$ 0.07} & \textbf{11.86 $\pm$ 0.27}

\end{tabular}
\vspace{-1.em}
\end{table}

\vspace{.8em}\noindent \textbf{Influence of the contrastive loss.} Table \ref{tab:multi_matching_results} presents the experimental results given by the three architectures: Siamese, Pseudo-Siamese and TS-Net network with/without the additional contrastive loss. In general, we observed that the error can be reduced by up to 30 \% by adding this loss, for any architecture and dataset. More importantly, this gain can be obtained with negligible computing costs and with little effort. During training, we found that the margin $Q$ and the weighting value $\lambda$ and $\beta$ are crucial and need to be carefully cross-validated. We also consider replacing it by the classical contrastive loss. However it turned out to be very sensitive to gradient explosion. In addition, to make these parameters less sensitive during training, we tried to normalize the features before feeding into the loss function in order to maintain a fixed range of distances. Unfortunately, we observed some (marginal) drop in performance. 

\vspace{.8em}\noindent \textbf{Comparison to Siamese and Pseudo-Siamese network.} Intuitively, the Pseudo-Siamese network has more parameters and degree of freedom to project the two modalities onto the new subspace. Hence, it should produce better results compared to the Siamese network (See Table \ref{tab:multi_matching_results}). However, in practice, we observed the opposite. We perform a grid search on the different parameters, regularization techniques (dropout, L2/L1), different losses (entropy/contrastive loss) with different strategy of combining the two towers (concatenation/subtraction). In all the experimentations, the Siamese network always outperform the Pseudo-Siamese network. This behavior has also been observed by \cite{zagoruyko2015learning, aguilera2016learning, merkle2017exploiting}. When combining the Siamese and Pseudo-Siamese network, we notice significant improvement over the 3 datasets. On VeDAI and CUHK, the error is reduced by almost 50\% not counting the additional loss at the feature level. On the three datasets, our approach outperforms the Siamese and Pseudo-Siamese networks. This fully justifies the competitiveness of our approach.
% \begin{table*}[]
% \centering
% \caption{The performance of the original MatchNet and the Re-implementation version. Not, Yos and Lib stand for Notredame, Yosemite and Liberty subset.}
% \label{tab:matchnet_reimplementation}
% \begin{tabular}{l|ll|ll|ll|l}
% Train set & Not  & Not   & Yos  & Yos   & Lib  & Lib   & Average \\ \hline
% Test set  & Lib  & Yos   & Not  & Lib   & Not  & Yos   &         \\
% MatchNet  & 9.48 & 12.17 & 8.27 & 15.40 & 5.18 & 14.40 & 10.82   \\
% Siamese+C &      &       &      &       &      &       &         \\
% TS-Net     & 8.90 &11.48  & 6.86 & 13.21 & 4.76 &12.70  & 9.654   \\
% TS-Net+C   &      &       &      &       &      &       &        
% \end{tabular}
% \end{table*}
% \noindent \textbf{Uni-modality matching}: 

\section{Conclusions}
\label{sec:conclusion}

We proposed a novel architecture, called TS-Net, for multimodal patch matching. TS-Net consists of two sub-networks: a Siamese and Pseudo-Siamese network. Each of them is responsible for learning different types of complementary characteristics from both modalities. In addition, we showed that an additional loss, at the intermediate feature level, is beneficial at the price of only a small additional computational costs. Experimental results demonstrate the superiority of our approach over Siamese and Pseudo-Siamese networks.

\vspace{0.8em}
\noindent \textbf{Acknowledgements.} This work was partly funded by the French--UK MCM ITP program and by the ANR-16-CE23-0006 program. The authors thank Shivang Agarwal for proofreading the manuscript.  
\bibliographystyle{IEEEbib}
\bibliography{strings,refs}

\end{document}